\documentclass[final]{cvpr}

\usepackage{times}
\usepackage{epsfig}
\usepackage{graphicx}
\usepackage{amsmath}
\usepackage{amssymb}

\DeclareMathOperator*{\argmax}{arg\,max}

\usepackage{multirow}
\usepackage{graphics}
\usepackage{xcolor}
\usepackage{hyperref}
\usepackage{caption}
\usepackage{subcaption}
\usepackage{algorithm}
\usepackage{algorithmic}
\usepackage{graphicx}

\usepackage{hyperref}
\hypersetup{pagebackref=true,breaklinks=true,colorlinks,bookmarks=false}

\def\cvprPaperID{} 
\def\confYear{CVPR 2021}

\begin{document}

\title{Rethinking the Spatial Route Prior in Vision-and-Language Navigation}

\author{Xinzhe Zhou$^1$, Wei Liu$^2$, Yadong Mu$^{1}$\footnote{Corresponding author.}\\
$^1$Peking University, $^2$Tencent Corp.}

\maketitle




\begin{abstract}
Vision-and-language navigation (VLN) is a trending topic which aims to navigate an intelligent agent to an expected position through natural language instructions. This work addresses the task of VLN from a previously-ignored aspect, namely the spatial route prior of the navigation scenes. A critically enabling innovation of this work is explicitly considering the spatial route prior under several different VLN settings. In a most information-rich case of knowing environment maps and admitting shortest-path prior, we observe that given an origin-destination node pair, the internal route can be uniquely determined. Thus, VLN can be effectively formulated as an ordinary classification problem over all possible destination nodes in the scenes. Furthermore, we relax it to other more general VLN settings, proposing a sequential-decision variant (by abandoning the shortest-path route prior) and an explore-and-exploit scheme (for addressing the case of not knowing the environment maps) that curates a compact and informative sub-graph to exploit. As reported by~\cite{ZhuQNSBWWEW21}, the performance of VLN methods has been stuck at a plateau in past two years. Even with increased model complexity, the state-of-the-art success rate on R2R validation-unseen set has stayed around 62\% for single-run and 73\% for beam-search. We have conducted comprehensive evaluations on both R2R and R4R, and surprisingly found that utilizing the spatial route priors may be the key of breaking above-mentioned performance ceiling. For example, on R2R validation-unseen set, when the number of discrete nodes explored is about 40, our single-model success rate reaches 73\%, and increases to 78\% if a Speaker model is ensembled, which significantly outstrips previous state-of-the-art VLN-BERT with 3 models ensembled.



\end{abstract}

\section{Introduction}\label{sec:intro}
Humans have a long-standing dream of creating intelligent robots that can accomplish different tasks following our instructions. Recent research about autonomous navigation has made tremendous advances towards this goal. In particular, the work of Anderson \emph{et.al.}~\cite{AndersonWTB0S0G18} attracted much attention as it involved both vision-language interaction and embodied navigation, known as Vision-and-Language Navigation (VLN). This task requires an autonomous agent to reach an expected goal position guided by purely natural language instructions. The release of several VLN benchmarks, including the Room-to-Room (R2R)~\cite{AndersonWTB0S0G18}, Room-for-Room~\cite{JainMKVIB19} (R4R), TOUCHDOWN~\cite{ChenSMSA19}, and Room-across-Room~\cite{KuAPIB20} (RxR) has spurred the rapid development of VLN models. The key performance metrics, such as success rate (SR), have witnessed continual improvement in the past years. Interestingly, a majority of existing works are formulated under the reinforcement learning (RL) paradigm. We are surprised to notice that most of previous works have ignored the spatial route prior inherent in the VLN task. 




This work aims to analyze various types of spatial route priors and harness them for enhancing the performance of VLN models. Our formulation starts by considering a naive case with shortest-path route prior and all navigation map known. As a direct derivation of the shortest-path prior, a pair of origin-destination would often uniquely determine the internal path that an agent goes through (the existence of multiple shortest paths for the same node pair is possible yet very rare). Therefore, under this strictly-constrained VLN setting, finding the best-matching route for an instruction boils down to choosing a node in the navigation map as the destination. We can thus frame the path-finding problem as an ordinary classification task over all valid nodes, which is more amenable to numerical optimization and tends to bring superior performances in comparison with conventional RL-based models as empirically verified in our experiments. The merits of the classification-based formulation are two-fold. First, RL-based methods suffer from an exponential number of candidate paths, while our proposed formulation reduces it to be linear to the number of nodes in the navigation scenes. For example, the R2R data has an average vertex degree of 4.1 and the ground-truth path contains 4 to 6 steps. Simple calculations reveal that RL-based methods would consider about 283 to 4,750 possible candidates, compared with 117 on average in our classification-based formulation. A second merit is classification on the whole-path level could make global inference about the route. RL-based agents instead generate the route from sequential local decisions which may lead to sub-optimal solutions. The issue of locality is only partly mitigated in recent works that integrate backtracking~\cite{KeLBHGLGCS19,MaWAXK19} or looking-ahead~\cite{WangWSLS20} mechanisms.

In realistic applications, the topology of navigation scenes is often unknown, and the shortest-path prior may not hold on a few challenging VLN benchmarks, such as R4R~\cite{JainMKVIB19} and RxR~\cite{KuAPIB20}. We thus further propose two variants of the basic classification-based VLN model. First, the shortest-path prior is relaxed to be more general, allowing a ground-truth path to be the composition of an arbitrary (usually small) number of local shortest sub-paths. Based on this relaxation, we adapt the classification model into a progress-aware sequential-decision variant, which resembles previous RL agents but classifies among the shortest sub-paths instead of all neighboring edges. Secondly, the assumption of knowing the topology of navigation maps is removed. We develop an informative partial topology (sub-graph) of the environment on-the-fly by re-purposing a weak Follower~\cite{FriedHCRAMBSKD18} model, upon which the (sequential) classification could be conducted.

It is important to point out that the spatial route prior is not just some dataset-dependent bias, but a general and natural property inherent in the VLN task. Intuitively, when humans are guiding the agent to some expected position, we are likely to choose the shortest route for efficiency. Or we may hope the agent to pass several places before reaching the goal, then a natural choice is to move along each local shortest sub-path between adjacent places sequentially. Interestingly, current VLN datasets reflect these different priors by construction. In R2R~\cite{AndersonWTB0S0G18}, the routes are all shortest paths connecting the starting and ending points, which corresponds to the former case. In more general R4R~\cite{JainMKVIB19} and RxR~\cite{KuAPIB20}, each route is likely to be composed of few shortest sub-paths connected end to end. To be specific, we used a greedy algorithm (see Algorithm~\ref{alg:decom}) to decompose each route and
calculated the $\frac{\# \text{decomposed shortest sub-paths}}{\# \text{original steps}}$ for all training routes in both datasets, and the values are about 0.21 and 0.25 on average. For comparison, we also calculated the above ratio for randomly-sampled routes in the same environments, and the result is about 0.79. The notable margin between R4R / RxR and random routes clearly demonstrates that the routes chosen in VLN have a strong prior consistent with the latter case above. Our main contribution in this work is thus to reveal these priors and integrate them in the model designs under different settings.

We validated the effectiveness of our method on both R2R and R4R datasets for the two different priors, respectively.
Our experiments clearly showed that by exploring only sub-graphs of moderate size, our model could outperform previous state-of-the-art methods with significant margins. For example, on the R2R validation-unseen set, when the number of discrete nodes explored is about 40, our single-model success rate reaches 73\%, and increases to 78\% if a Speaker~\cite{FriedHCRAMBSKD18} model is ensembled, which even out-performs the 3-model ensembled VLN-BERT~\cite{MajumdarSLAPB20}. The latter even utilizes large amount of extra data for pre-training while we need not. Our evaluations have provided strong evidence that the spatial route priors may be the key for breaking the performance ceilings as reported by recent VLN works~\cite{ZhuQNSBWWEW21}.

\section{Related Work}


In an early VLN work~\cite{AndersonWTB0S0G18}, Anderson \emph{et.al.} designed a simple sequence-to-sequence (Seq2Seq) agent that operated directly in low-level action spaces and was trained by purely Imitation-Learning (IL). This model set a baseline success-rate (SR) that was much lower than human performance (20\% v.s. 86\%). After it, numerous algorithms were continually developed for improvement in different aspects. For instance, \cite{WangXWW18} changed the pure-IL training to a hybrid of model-based and model-free Reinforcement-Learning (RL). \cite{WangHcGSWWZ19} further designed an ensemble of IL and RL training paradigm, which has now become the default choice. The work of \cite{FriedHCRAMBSKD18} revised the original low-level action space to a more high-level panoramic space, which greatly reduced the sequence length and was followed by almost all later works. Besides, \cite{FriedHCRAMBSKD18} also proposed a dual Speaker-Follower framework, which helped in both data augmentation and inference-time ensemble. Some later works attempted to introduce heuristic mechanism into the agent like backtracking / regret~\cite{KeLBHGLGCS19,MaWAXK19}, active exploration~\cite{WangWSLS20}. Another line of research was trying to improve the unseen generalization through either pre-training~\cite{HuangJMKMBI19,HaoLLCG20,MajumdarSLAPB20}, auxiliary supervision~\cite{MaLWAKSX19,WangJIWKR20,Zhu0CL20,WangWS20}, or training-data processing~\cite{TanYB19,FuWPGEW20,ParvanehATSH20}. The basic Seq2Seq structure had also been improved by introducing cross-modal attention~\cite{WangHcGSWWZ19} and fine-grained relationship~\cite{HongOQ0G20}, utilizing the semantic or syntactic information of languages~\cite{QiPZHW20,LiTB21}, reformulating under a Bayesian framework~\cite{AndersonSPBL19}, and combining long-range memory for global decision~\cite{DengNR20,WangWLXS21}. Besides R2R, some later works also proposed more challenging datasets like Room-for-Room (R4R)~\cite{JainMKVIB19}, TOUCHDOWN~\cite{ChenSMSA19}, and Room-across-Room~\cite{KuAPIB20} (RxR).

\section{Our Approach}\label{sec:method}
As stated in Section~\ref{sec:intro}, mainly two kinds of route priors are observed in VLN benchmarks. For clarity, we would first describe our method for both of them under an information-rich setting (Sec.~\ref{ssec:spp} and \ref{ssec:relax}), then we explain how we generalize it to the realistic scenario as in current benchmarks (Sec.~\ref{ssec:subgraph}).


\subsection{Naive Case: Shortest-Path Prior + Known Map}
\label{ssec:spp}


Assume that a navigation environment can be represented by an edge-weighted graph $\mathcal{G} = \{\mathcal{V}, \mathcal{E}\}$, where $\mathcal{V}$ collects all valid observation positions and $\mathcal{E}$ covers all edges between neighboring nodes. In a typical setting of VLN, an agent is initially located at some
$n_s\in \mathcal{V}$ and moves once receiving an $l$-word navigation instruction $\mathcal{I} = (w_1, w_2,\dots, w_l)$. For the instruction $\mathcal{I}$, there always exists a ground-truth path $\mathcal{P}_{n_s, gt_t}=(n_{gt_1},n_{gt_2},\dots,n_{gt_t})$, where $n_{gt_1}=n_s$ and $(n_{gt_j}, n_{gt_{j+1}}) \in \mathcal{E}$ for any step $j$. $t$ is the length of the path and often varies over different instructions. 

Let us first consider a naive case: the full information of graph $\mathcal{G}$ is known and each ground-truth path is the shortest path between the origin-destination pair, namely the agent operates with a shortest-path prior and known map. Such a setting paves a way to learn an effective albeit simple VLN model which is essentially different from popular RL-based solutions. In specific, given an instruction $\mathcal{I}$ and a starting node $n_s$, the shortest path $\mathcal{P}_{n_s, n_i}$ between $n_s$ and $n_i$ can be uniquely determined via standard Floyd or Dijkstra algorithm~\cite{Dijkstra59}. This fact critically leads to classification-based path-finding solver conditioned on $\mathcal{I}$ and $n_s$:
\begin{equation}
    \argmax_{n_i} \mathcal{F}(\mathcal{P}_{n_s,n_i}, \mathcal{I}), ~~\forall \ n_i\in \mathcal{V}\label{eqn:1},
\end{equation}
where $\mathcal{F}(\cdot)$ stands for any well-defined path-instruction scoring function.

One of the ambitions in this paper is to demonstrate that the simple classification in Eqn.~\ref{eqn:1}, compared with various sophisticated RL-based policies in existing work, can serve as a major innovation that significantly advances the state-of-the-art of VLN. To this end, we directly adopt a naive off-the-shelf dual-LSTM~\cite{HochreiterS97} based model to implement $\mathcal{F}$, which has been widely adopted in previous practice~\cite{FriedHCRAMBSKD18,MaLWAKSX19,TanYB19,Zhu0CL20,WangWSLS20}. In detail, 
two separate LSTMs are used to encode the path and the instruction, denoted as $\text{LSTM}_\text{path}$, $\text{LSTM}_\text{instr}$ respectively. Their final hidden-state vectors are concatenated and fed into an MLP, obtaining a compatibility-indicating score. The optimization of this model can be guided by an ordinary cross-entropy loss calculated on exhaustively-sampled path-instruction pairs on the training navigation scenes, where the ground-truth pairs are known.

One intriguing property of the VLN task is the judgment of a successful navigation. A commonly-adopted metric, success rate (SR)~\cite{AndersonWTB0S0G18}, regards any navigational path that ends at any position within specific distance (a user-settable hyper-parameter, say 3 meters) of the ground-truth goal to be correct. This subtly diverges from standard classification problems with one-hot ground-truth labels. To fully harness such property for enhancing the navigation success rate, we stick to the ordinary Softmax neural layer for generating activation scores, yet at the inference time, a \emph{neighboring score aggression} scheme is proposed to compensate those sufficiently close yet not exact solutions. Namely, the optimal destination is calculated via:
\begin{equation}
    \argmax_{n_j} ~\text{sum}_{n_j=n_i ~\text{or}~ n_j\in \mathcal{N}(n_i)} \{ \mathcal{F}(\mathcal{P}_{n_s,n_j}, \mathcal{I}) \}, ~~\forall \ n_j\in \mathcal{V}\label{eqn:2},
\end{equation}
where the index set $\mathcal{N}(n_i)$ contains all nodes within a small distance (3 meters in our experiments) of the node $n_i$, and $\text{sum}_{n_j\in \mathcal{N}(n_i)}$ defines an operator that sums all scores from $\mathcal{N}(n_i)$. More details about hyperparameters and training schedules are summarized in the supplementary material.


\begin{algorithm}[t!]
  \caption{Decompose a route into sequential shortest sub-paths}
  \label{alg:decom}
  \begin{small}
\begin{algorithmic}[1]
  \STATE {\bfseries Input:} A route $\mathcal{P}=(n_1, n_2, \dots, n_t)$ and the corresponding graph $\mathcal{G} = \{\mathcal{V}, \mathcal{E}\}$;
  \STATE {\bfseries Output:} The decomposed shortest-paths $\mathcal{P}^\prime$;
  \vspace{0.04in}
  \STATE Compute the shortest-paths between any two nodes in $\mathcal{G}$; Initialize $\mathcal{P}^\prime$ to an empty list and $\textit{i}$ to 1;
  \WHILE{$i \leq t$}
  \STATE $\textit{j} \gets i$;
  \WHILE{$j \leq t$ and $(n_i, n_{i+1},\dots,n_j)$ is the shortest-path between $n_i$ and $n_j$}
  \STATE $\textit{j} \gets j+1$;
  \ENDWHILE
  \STATE Append $(n_i, n_{i+1}, \dots, n_{j-1})$ to the tail of $\mathcal{P}^\prime$ and mark $(n_i,n_{j-1})$ as a local starting-ending node pair;
  \STATE $i \gets j-1$;
  \ENDWHILE
  \vspace{0.01in} \newline
\end{algorithmic}
\end{small}
\vspace{-0.15in}
\end{algorithm}

\begin{figure*}[t!]
  \centering
  \includegraphics[width=\linewidth]{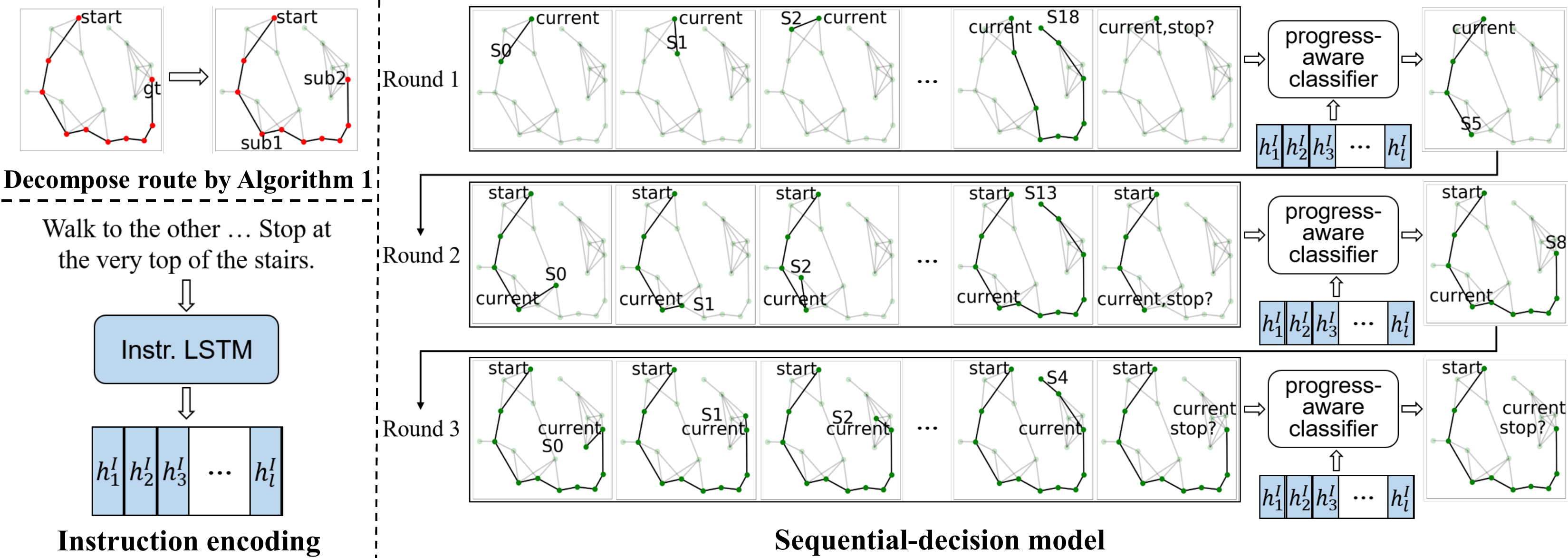}
  \caption{\small An example illustrating the route decomposition and the sequential-decision process. In the upper part of the left panel, we show an example route and its decomposition after Algorithm~\ref{alg:decom}. The lower part shows the instruction-encoding process. In the right panel, a three-round sequential-decision process is shown. The agent starts from ``current" and chooses the shortest path to ``S5" as the first sub-path in the 1st round. In the 2nd round, the agent expands the route to ``S8" with a new shortest path. The route-finding ends properly after the 3rd round when the agent chooses the ``Stop" option. The whole route is finally formed through all the three rounds.}
  \label{fig:r4r}
\end{figure*}

\subsection{Relaxation to General Route Prior}
\label{ssec:relax}

As stated before, the shortest-path prior does not always hold on all VLN benchmarks and realistic scenarios, such as the R4R~\cite{JainMKVIB19} and RxR~\cite{KuAPIB20} datasets. To extend the idea in Sec.~\ref{ssec:spp} to more general cases, this section elaborates on a second scheme that addresses a setting with the more general spatial route prior: each path is composed of several (usually few) sub-paths connected end to end, each of which admits a local shortest path with respect to its starting and ending nodes. Our proposed method starts from decomposing an arbitrary ground-truth route into consecutive shortest sub-paths. Note that the desired decomposition may not be unique and multiple valid ones may exist. In the VLN task, a decomposition with minimal number of sub-paths is favored, since it will require fewer route-finding operations and largely match humans' nature. To this end, we propose a one-pass process that greedily pursues a local shortest sub-path with maximal steps. The pseudo-code for the process is found in Algorithm~\ref{alg:decom}.

Given an instruction $\mathcal{I}$ and the corresponding ground-truth path $\mathcal{P}$, note that the output of Algorithm~\ref{alg:decom} is a list of starting-ending node pairs. Denote them as $(n_s^{(1)},n_e^{(1)}),(n_s^{(2)},n_e^{(2)}),\ldots,(n_s^{(K)},n_e^{(K)})$, where $K$ counts the sub-paths return by Algorithm~\ref{alg:decom}. This inspires us to tailor the method described in Sec.~\ref{ssec:spp} to a sequential-decision variant. Concretely, the agent attempts to accomplish a route-finding task in $K$ rounds. At round $k$, the model is optimized to navigate the agent from $n_s^{(k)}$ to $n_e^{(k)}$, ideally forming a partial path $n_s^{(1)} \rightarrow n_e^{(k)}$ with $k$ shortest sub-paths. We formulate each round as a classification problem as in Sec.~\ref{ssec:spp}, with $n_s^{(k)}$, $n_e^{(k)}$ to be the starting node and locally ground-truth destination node respectively. Such iterative decisions progressively expand the partial path, targeting the ground-truth path $\mathcal{P}$. As the total number of rounds $K$ is unknown for the agent, we introduce a special learnable candidate to indicate the ``Stop" action, which is jointly optimized through the classification process in each round. An example from R4R illustrating the above process is shown in Figure~\ref{fig:r4r}.

The function $\mathcal{F}(\cdot)$ in Eqn.~(\ref{eqn:1}) gauges the path-instruction compatibility. However, at each round of above sequential-decision process, $\mathcal{F}(\cdot)$ will be fed with a full instruction yet partial path, leading to imprecise scores. To remedy this, we propose a \emph{progress-aware scoring function}. At specific round, suppose the current pursued path is $(n_1,n_2,\ldots,n_{\tilde t})$, with $\tilde t$ nodes. Note $n_1$ is the original starting node, \emph{i.e.} we always encode the whole path history for each intermediate sub-path selection. Denote the dual-LSTM outputs as
\begin{eqnarray}
    h^I_1, h^I_2, \dots, h^I_l &=& \text{LSTM}_{\text{instr}}(w_1, w_2, \dots, w_l)\label{eqn:3},\\
    h^P_1, h^P_2, \dots, h^P_{\tilde{t}} &=& \text{LSTM}_{\text{path}}(n_1, n_2, \dots, n_{\tilde{t}})\label{eqn:4},
\end{eqnarray}
where $h^I_j$ and $h^P_j$ represent the $j$-th step hidden states of the encoded instruction and path, respectively. 

The main fact that we harness here is that $h^P_{\tilde{t}}$ is likely to match some one in $\{h^I_1, h^I_2, \dots, h^I_l\}$ if the current round successfully finds the correct partial path. Inspired by soft-DTW~\cite{CuturiB17}, we fetch the hidden state at every step of the instruction and concatenate each of them with the path final state to compute the scores corresponding to different progresses / steps using a similar MLP as in Sec.~\ref{ssec:spp}, according to
\begin{eqnarray}
    s_j &=& \text{MLP}([h^I_j, h^P_{\tilde{t}}]),~~\forall~j = 1,2,\dots,l \label{eqn:5}.
\end{eqnarray}



Then, we conduct an aggregation to obtain a final score:
\begin{eqnarray}
    s &=& \text{Aggr}_\text{soft}\left(s_1,s_2,\dots,s_l \right)\nonumber\\
    &=& \text{log} \left(\text{exp}(s_1) + \text{exp}(s_2) + \dots + \text{exp}(s_l) \right).\label{eqn:7}
\end{eqnarray}

Simply put, 
$\text{Aggr}_\text{soft}$ is a soft maximum operation which behaves like soft-attention~\cite{LuongPM15} to aggregate different progresses. 
The comparison of $\text{Aggr}_\text{soft}$ and solely the MLP function is shown in Section~\ref{ssec:ablate}.


\begin{figure*}[t!]
  \centering
  \includegraphics[width=0.85\linewidth]{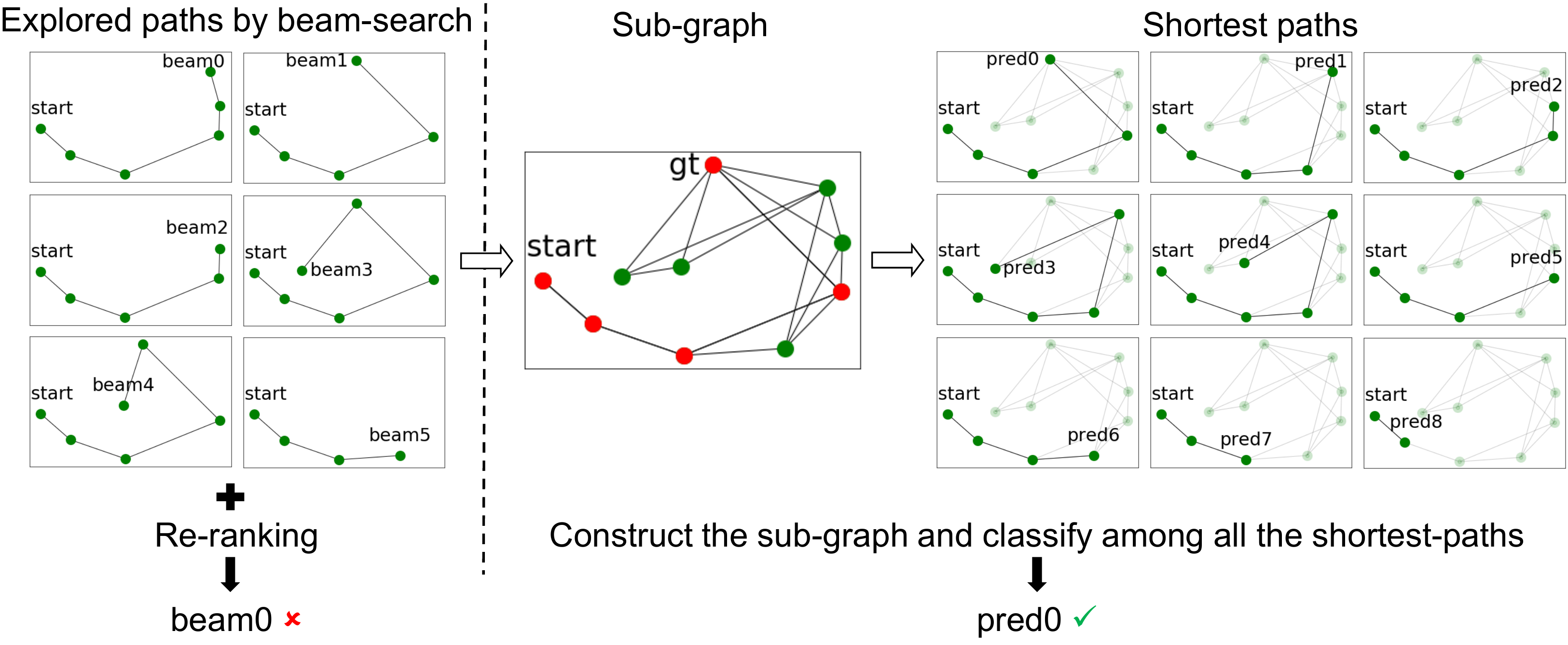}
  \caption{\small An example illustrating our explore-and-exploit scheme (right) with comparison to previous beam-search and re-ranking paradigm (left) from R2R.}
  \label{fig:compare}
\end{figure*}

The above model is akin to existing RL-based agents~\cite{FriedHCRAMBSKD18,TanYB19} in that both operate sequentially and form the whole route from sub-paths. Nonetheless, there are several important differences between them. First, each decision in previous RL-based agents chooses one more edge in the route, leading to many steps to return the ground-truth route. In contrast, each decision in our model will include a longer sub-path, thus greatly reducing the number of required decisions. Take R4R as an example, with previous agents, the average length of the decision sequence is about 12, while with our decomposition as Algorithm~\ref{alg:decom}, the average number of the sub-paths is only 3.3. Besides, our sequential model inherits the global-view merit of the classification model in Section~\ref{ssec:spp}. Compared with previous methods that make local decisions in the instant neighbors, our model classifies among the shortest-paths from current node which extends the view to several steps away and enables a more global decision. This helps alleviate the chance that the model falls into local minimum (w.r.t. the loss).

The optimization of the above model follows the previous practice~\cite{WangHcGSWWZ19}, by jointly using Imitation learning (IL) and RL. For IL, we use teacher-forcing and feed the teacher choice for each decision by Algorithm~\ref{alg:decom}. For RL, we use the CLS-based reward proposed in~\cite{JainMKVIB19}, as it encourages to reach not only the goal but the fidelity to the ground-truth route. More details about the training are included in the supplementary material.

\subsection{Relaxation to Unknown Map}
\label{ssec:subgraph}

Both of above models operate well when the environment topology is known beforehand, such that all shortest-paths could be exactly computed. It is more realistic to assume the environment is totally unseen before the navigation, which is also the standard setting in current benchmarks. To generalize our model to such setting, we here develop an explore-and-exploit scheme. It works as follows: first, after receiving each instruction,
we run another RL agent in the state-factored search (a.k.a. beam-search) manner proposed by \cite{FriedHCRAMBSKD18}, in order to construct an informative sub-graph of the environment. The policy can be any off-the-shelf one reported in existing VLN works, and we choose a simple Follower~\cite{FriedHCRAMBSKD18} model in our main experiments. The sub-graph is formed by all the visited nodes in the environment, and is informative since agents like Follower could perceive the instruction and try to include the ground-truth route in the sub-graph, although it is not effective enough to identify it.  

Next, according to the route prior, our classification model or the sequential-decision variant is utilized. The model runs on the sub-graph found in the first step, returning the path with highest path-instruction score, and derives the agent along the predicted route.

We adopt such explore-and-exploit scheme for benchmarking our method in standard datasets like R2R and R4R. Note our method could not be directly compared with most previous single-run models, as an extra exploration phase is involved in our framework. However,
our explore-and-exploit scheme is similar to the popular beam-search and re-ranking paradigm~\cite{FriedHCRAMBSKD18,MajumdarSLAPB20}, in the sense that both need to explore the environments before the final decision. The only difference is that we construct a sub-graph from all the explored nodes and exploit the route prior over the sub-graph. While for re-ranking methods, they collect the candidates (beams) and only choose one best from them, ignoring the graph topology and route priors. An example illustrating the differences is shown in Figure~\ref{fig:compare}. Therefore in the experiments, we mainly compare with previous beam-search methods.



\section{Experiment}\label{sec:exp}
\subsection{Results on R2R Dataset}\label{ssec:r2r}
\textbf{Dataset.}
Room-to-Room (R2R)~\cite{AndersonWTB0S0G18} is among the first and most widely adopted datasets for benchmarking VLN performance. It contains rich panoramic scenes of 90 different houses / environments with manually annotated instruction-route pairs. The dataset is split into \texttt{train} (61 environments, 14,025 instructions), \texttt{val-seen} (56 environments, 1,020 instructions), \texttt{val-unseen} (11 environments, 2,349 instructions), and \texttt{test-unseen} (18 environments, 4,173) sets. 

\textbf{Evaluation Metric.}
Following \cite{AndersonWTB0S0G18,abs-1807-06757}, we use five metrics: (1) \emph{Success Rate} (SR) counts for the ratio of the agent stopping at any position within 3 meters of the ground-truth goal, and is taken as the primary metric. (2) \emph{Navigation Error} (NE) quantitatively characterizes the error between the final position and the goal by their shortest-path distance (in meters). (3) \emph{Trajectory Length} (TL) computes the total length (in meters) of the agent's trajectory. (4) \emph{Oracle Success Rate} (OSR) is an upper-bound for SR computed with an oracle stopping-rule. (5) \emph{Success rate weighted by Path Length} (SPL) discounts the SR by considering the trajectory length.

\textbf{Our Evaluated Models.} As the shortest-path prior holds on R2R, we evaluate the classification model from Section~\ref{ssec:spp} with both map known and unknown. Since most previous methods do not assume known maps, our comparison is mainly based on the unknown-map model, and only include the known-map model as an upper-bound.  

  
\begin{table*}[t!]
  \caption{Results of different methods on R2R dataset.
  All methods are trained with the augmented data from back-translation. "-" means number not available. +spk means conducting model ensemble with a Speaker~\cite{FriedHCRAMBSKD18} model. +flw means conducting model ensemble with a Follower~\cite{FriedHCRAMBSKD18} model.}
  \label{tab:r2r}
  \begin{scriptsize}
  \begin{tabular}{c|ccccc|ccccc|ccccc}
    \hline
    \multirow{3}{*}{Models} & \multicolumn{15}{c}{R2R} \\
    & \multicolumn{5}{c|}{\texttt{val-seen}} & \multicolumn{5}{c}{\texttt{val-unseen}} & \multicolumn{5}{c}{\texttt{test-unseen}} \\ 
    & \textbf{SR}$\uparrow$ & NE$\downarrow$ & TL$\downarrow$ & OSR$\uparrow$ & SPL$\uparrow$ & \textbf{SR}$\uparrow$ & NE$\downarrow$ & TL$\downarrow$ & OSR$\uparrow$ & SPL$\uparrow$ & \textbf{SR}$\uparrow$ & NE$\downarrow$ & TL$\downarrow$ & OSR$\uparrow$ & SPL$\uparrow$ \\ \hline \hline
    \multicolumn{16}{c}{beam-search} \\ \hline
    Speaker-Follower~\cite{FriedHCRAMBSKD18} & 0.70 & 3.08 & - & 0.78 & - & 0.55 & 4.83 & - & 0.65 & - & 0.54 & 4.87 & 11.6 & 0.64 & - \\
    RCM~\cite{WangHcGSWWZ19} & - & - & - & - & - & - & - & - & - & - & 0.63 & 4.03 & - & - & - \\
    Self-Monitoring~\cite{MaLWAKSX19} & 0.71 & 3.04 & - & 0.78 & 0.67 & 0.58 & 4.62 & - & 0.68 & 0.52 & 0.61 & 4.48 & - & 0.70 & 0.56 \\ 
    Tactical Rewind~\cite{KeLBHGLGCS19} & 0.70 & 3.13 & - & - & - & 0.63 & 4.03 & - & - & - & 0.61 & 4.29 & - & - & - \\
    EnvDrop~\cite{TanYB19} & 0.76 & 2.52 & - & - & - & 0.69 & 3.08 & - & - & - & 0.69 & 3.26 & - & - & - \\
    AuxRN~\cite{Zhu0CL20} & - & - & - & - & - & - & - & - & - & - & 0.71 & 3.24 & - & - & - \\
    Active Perception~\cite{WangWSLS20} & - & - & - & - & - & - & - & - & - & - & 0.70 & 3.07 & - & - & - \\
    VLN-BERT~\cite{MajumdarSLAPB20} & 0.70 & 3.73 & 10.3 & 0.76 & 0.66 & 0.59 & 4.10 & 9.6 & 0.69 & 0.55 & - & - & - & - & -  \\
    VLN-BERT~\cite{MajumdarSLAPB20}+spk+flw & 0.82 & 2.35 & 10.6 & 0.87 & 0.78 & 0.74 & 2.76 & 10.0 & 0.82 & 0.68 & 0.73 & 3.09 & - & - & - \\

    \textbf{Ours(explore@40)} & 0.74 & 3.72 & 9.6 & 0.77 & 0.71 & 0.73 & 3.61 & 9.0 & 0.78 & 0.70 & 0.71 & 3.81 & 9.3 & 0.75 & 0.69 \\
     
    \textbf{Ours(explore@40)}+spk & 0.78 & 3.27 & 9.7 & 0.81 & 0.75 & 0.78 & 3.13 & 9.1 & 0.83 & 0.74 & 0.74 & 3.55 & 9.4 & 0.78 & 0.72  \\ 
    \hline \hline
    
    \textbf{Ours(known map)} & 0.76 & 3.69 & 9.7 & 0.79 & 0.72 & 0.74 & 3.56 & 9.1 & 0.80 & 0.71 & 0.72 & 3.77 & 9.4 & 0.76 & 0.69 \\
    
    \textbf{Ours(known map)}+spk & 0.80 & 3.20 & 9.8 & 0.83 & 0.77 & 0.79 & 3.05 & 9.3 & 0.85 & 0.76 & 0.76 & 3.47 & 9.5 & 0.79 & 0.73 \\
     \hline
 \end{tabular}
 \end{scriptsize}
\end{table*}

\begin{table}[t!]
  \centering
  \caption{R2R online leaderboard results.  "-" means number not available. +spk means conducting model ensemble with a Speaker~\cite{FriedHCRAMBSKD18} model. +flw means conducting model ensemble with a Follower~\cite{FriedHCRAMBSKD18} model. }
  \label{tab:r2r_submit}
  \centering
  \begin{footnotesize}
  \begin{tabular}{c|ccccc}
    \hline
    \multirow{3}{*}{Models} & \multicolumn{5}{c}{R2R} \\
    & \multicolumn{5}{c}{\texttt{test-unseen}} \\ 
    & \textbf{SR}$\uparrow$ & NE$\downarrow$ & TL$\downarrow$ & OSR$\uparrow$ & SPL$\uparrow$ \\ \hline \hline
    \multicolumn{5}{c}{beam-search} \\ \hline
    Speaker-Follower~\cite{FriedHCRAMBSKD18} & 0.54 & 4.87 & 1257.4 & 0.96 & 0.01 \\
    RCM~\cite{WangHcGSWWZ19} & 0.63 & 4.03 & 357.6 & 0.96 & 0.02 \\
    Self-Monitoring~\cite{MaLWAKSX19} & 0.61 & 4.48 & 373.1 & 0.97 & 0.02\\ 
    Tactical Rewind~\cite{KeLBHGLGCS19} & 0.61 & 4.29 & 196.5 & 0.90 & 0.03\\
    EnvDrop~\cite{TanYB19} & 0.69 & 3.26 & 686.8 & 0.99 & 0.01\\
    AuxRN~\cite{Zhu0CL20} & 0.71 & 3.24 & 40.8 & 0.81 & 0.21\\
    Active Perception~\cite{WangWSLS20} & 0.70 & 3.07 & 176.2 & 0.94 & 0.05\\
    VLN-BERT~\cite{MajumdarSLAPB20}+spk+flw & 0.73 & 3.09 & 686.6 & 0.99 & 0.01 \\ 
    
    \textbf{Ours(explore@40)} & 0.71 & 3.81 & 625.1 & 0.99 & 0.01 \\
    \textbf{Ours(explore@40)}+spk & 0.74 & 3.55 & 625.3 & 0.99 & 0.01 \\ 
    \hline \hline
    
    \textbf{Ours(known map)} & 0.72 & 3.77 & 533.6 & 1.00 & 0.02 \\
    \textbf{Ours(known map)}+spk & 0.76 & 3.47 & 533.7 & 1.00 & 0.02 \\
    
    \hline
    
 \end{tabular}
  \end{footnotesize}
\end{table}

\textbf{Quantitative Result and Analysis.}
The results of our method with and without the map known on R2R are shown in Table~\ref{tab:r2r}, together with other competing approaches under the beam-search setting. Note the exploration trajectories are not taken into account in Table~\ref{tab:r2r} to compare solely the path-finding ability of different methods. The full metrics accounting the exploration phase on \texttt{test-unseen} are shown in Table~\ref{tab:r2r_submit}.
By default, our method explores up to 40 nodes in the environment when the map is unknown. Several conclusions could be drawn from Table~\ref{tab:r2r}. 

First, regarding the primary metric SR, our method shows great superiority on both \texttt{val-unseen} and \texttt{test-unseen}, and beats most competitors on \texttt{val-seen}. 
Particularly, when combined with a sole Speaker~\cite{FriedHCRAMBSKD18} model, our model could even out-perform the three-model ensembled VLN-BERT~\cite{MajumdarSLAPB20} which needs pretraining on large amount of data, demonstrating the great effectiveness of harnessing the route prior. 
Furthermore, comparing the results on \texttt{val-seen} and \texttt{val-unseen} / \texttt{test-unseen}, most methods except for ours show large performance drop when tested on unseen environments, which implies that they are prone to overfitting to the training scenes to promote the results on \texttt{val-seen}. Instead, our model exhibits similar performance on all three sets, demonstrating its good generalization ability.
Besides, the TL of our methods are shorter than others, thanks to the shortest-path prior we utilize, which also contributes to the high SPL. 

However, we also notice that the our NEs are worse than several methods despite higher SR. We believe this is one drawback of the classification model: an erroneous classification could induce large deviation from the goal. 
This is different from previous RL-based agents, where one decision only moves the agent one step, so even when the route is deviated by some choice, the agent could still navigate to the goal through later rectifications. 

Lastly, comparing among all our methods, it can be seen that utilizing the known map is almost always better than exploring a sub-graph. There are mainly two reasons: (1) With the full map, the ground-truth route is assured to be in the candidates, while may be out of the span of the explored sub-graph. (2) The neighboring score aggregation functions only approximately on the sub-graph since it is incomplete, and we show in Section~\ref{ssec:ablate} that it may even hamper the performance if the sub-graph is too small. 

\textbf{Leaderboard Submission Results}
For fairly benchmarking our method, we follow the previous conventions~\cite{FriedHCRAMBSKD18,MajumdarSLAPB20} to submit our results to the online server\footnote{\url{https://eval.ai/web/challenges/challenge-page/97/leaderboard}}. 
In Table~\ref{tab:r2r_submit}, we include all necessary trajectories for exploration in the submission. For the known-map model, we run a Depth-First Search (DFS) to explore the whole graph and return back to the starting point. For the sub-graph based model, we just append the trajectory of the exploration agent (and returning to the start) at the beginning.

The main difference of Table~\ref{tab:r2r_submit} with \ref{tab:r2r} is that the TLs now are all extremely long. But we point out that our methods achieve the best SR with similar TL as others, even smaller than VLN-BERT~\cite{MajumdarSLAPB20}. Note VLN-BERT utilized the same Follower~\cite{FriedHCRAMBSKD18} as ours for exploration, which demonstrates that the better effectiveness of our method is not due to improved exploration, but the proper utilization of the spatial route prior.

Besides, we find that the TL of exploring the full map by DFS is even smaller than exploring the sub-graph of size 40. The latter conducts the State-Factored Search (SFF)~\cite{FriedHCRAMBSKD18} which involves abundant transfer within the already-explored graph. This implies that for environments with moderate size like those in R2R, conducting simple DFS may be more preferred. But with larger environments, SFF should be a better choice as it maintains a relatively stable TL.

\subsection{Results on R4R Dataset}\label{ssec:r4r}
\textbf{Dataset.}
Room-for-Room (R4R)~\cite{JainMKVIB19} is an extended dataset based on R2R. It connects a pair of paths head-by-tail (with necessary connection path) and joins their instructions to form a long instruction-path pair, which breaks the shortest-path prior and challenges more for fidelity to the instruction. R4R contains three splits: \texttt{train} (61 environments, 233,613 instructions), \texttt{val-seen} (56 environments, 1,035 instructions), and \texttt{val-unseen} (11 environments, 45,162 instructions) set. The annotation of the test set of R2R is kept private so is not included.

\textbf{Evaluation Metric.}
Following \cite{JainMKVIB19,IlharcoJKIB19,DengNR20,WangWLXS21}, we keep the SR, TL, and NE from R2R, and introduce three new metrics: (1) \emph{Coverage weighted by Length Score} (CLS) measures the coverage of the model trajectory w.r.t. the ground-truth path. (2) \emph{normalized
Dynamic Time Warping} (nDTW) takes order-consistency into consideration, and (3) \emph{Success
rate weighted normalized Dynamic Time Warping} (SDTW) multiplies the SR with nDTW. The primary metric is CLS now.

\textbf{Our Evaluated Models.} For R4R, the shortest-path prior is no longer valid, so we only assume the general route prior and evaluate the sequential-decision model from Section~\ref{ssec:relax} with both map known and unknown. 
Since we haven't found any previous work conducting beam-search on R4R, we could not make direct comparisons with others. To try to provide a context for our performance, we additionally list several previous single-run methods, and compare them with solely our path-finding agent (the 2nd stage). Note this comparison is essentially not fair as we conducted extra exploration, so could only serve as a reference but not evidence.

\begin{table*}[t!]
  \centering
  \caption{Results of different methods on R4R dataset. "-" means number not available. }
  \label{tab:r4r}
  \centering
  \begin{footnotesize}
  \begin{tabular}{c|cccccc|cccccc}
    \hline
    \multirow{3}{*}{Models} & \multicolumn{12}{c}{R4R} \\
    & \multicolumn{6}{c|}{\texttt{val-seen}} & \multicolumn{6}{c}{\texttt{val-unseen}} \\ 
    & SR$\uparrow$ & NE$\downarrow$ & TL & \textbf{CLS}$\uparrow$ & nDTW$\uparrow$ & SDTW$\uparrow$ & SR$\uparrow$ & NE$\downarrow$ & TL & \textbf{CLS}$\uparrow$ & nDTW$\uparrow$ & SDTW$\uparrow$ \\ \hline 
    \multicolumn{13}{c}{single-run} \\ \hline
    Speaker-Follower~\cite{FriedHCRAMBSKD18} & 0.52 & 5.35 & 15.4 & 0.46 & - & - & 0.24 & 8.47 & 19.9 & 0.30 & - & - \\
    RCM(goal oriented)~\cite{JainMKVIB19} & 0.56 & 5.11 & 24.5 & 0.40 & - & - & 0.29 & 8.45 & 32.5 & 0.20 & 0.27 & 0.11 \\
    RCM(fidelity oriented)~\cite{JainMKVIB19} & 0.53 & 5.37 & 18.8 & 0.55 & - & - & 0.26 & 8.08 & 28.5 & 0.35 & 0.30 & 0.13 \\
    PTA(low-level)~\cite{abs-1911-12377} & 0.57 & 5.11 & 11.9 & 0.52 & 0.42 & 0.29 & 0.27 & 8.19 & 10.2 & 0.35 & 0.20 & 0.08 \\
    PTA(high-level)~\cite{abs-1911-12377} & 0.58 & 4.54 & 16.5 & 0.60 & 0.58 & 0.41 & 0.24 & 8.25 & 17.7 & 0.37 & 0.32 & 0.10 \\
    EGP~\cite{DengNR20} & - & - & - & - & - & - & 0.30 & 8.00 & 18.3 & 0.44 & 0.37 & 0.18 \\
    EnvDrop~\cite{TanYB19} & 0.52 & - & 19.9 & 0.53 & - & 0.27 & 0.29 & - & 27.0 & 0.34 & - & 0.09 \\
    OAAM~\cite{QiPZHW20} & 0.56 & - & 11.8 & 0.54 & - & 0.32 & 0.31 & - & 13.8 & 0.40 & - & 0.11 \\
    SSM~\cite{WangWLXS21} & 0.63 & 4.60 & 19.4 & 0.65 & 0.56 & 0.44 & 0.32 & 8.27 & 22.1 & 0.53 & 0.39 & 0.19 \\ \hline
    \multicolumn{13}{c}{beam-search} \\ \hline
    
    \textbf{Ours(explore@40)} & 0.45 & 6.35 & 19.8 & 0.66 & 0.58 & 0.36 & 0.36 & 7.02 & 20.2 & 0.60 & 0.50 & 0.27 \\
    
    \textbf{Ours(known map)} & 0.45 & 6.92 & 20.1 & 0.67 & 0.57 & 0.38 & 0.36 & 7.46 & 20.5 & 0.61 & 0.50 & 0.28 \\
    
    \hline
 \end{tabular}%
 \end{footnotesize}
\end{table*}

\textbf{Quantitative Result and Analysis.}
The results on R4R dataset are shown in Table~\ref{tab:r4r}.  
Firstly, our methods show good CLS and nDTW (SDTW) surpassing the state-of-the-art single-run models, which demonstrates the great fidelity to instructions. Regarding SR, although our model performs shyly on \texttt{val-seen}, it beats all other methods on \texttt{val-unseen} with clear margins. Similar to R2R, the large gap of other methods between \texttt{val-seen} and \texttt{val-unseen} shows that they overfit severely to the seen environments, while ours are more overfitting-free and rely mostly on the generalization ability. 
Lastly, comparing between our models, the difference is subtle with the known-map model slightly better, which is consistent with R2R.

Additional experiment results on the recent RxR~\cite{KuAPIB20} is included in the supplementary material with similar conclusions drawn.

\begin{figure}[t!]
     \centering
     \includegraphics[width=0.85\linewidth]{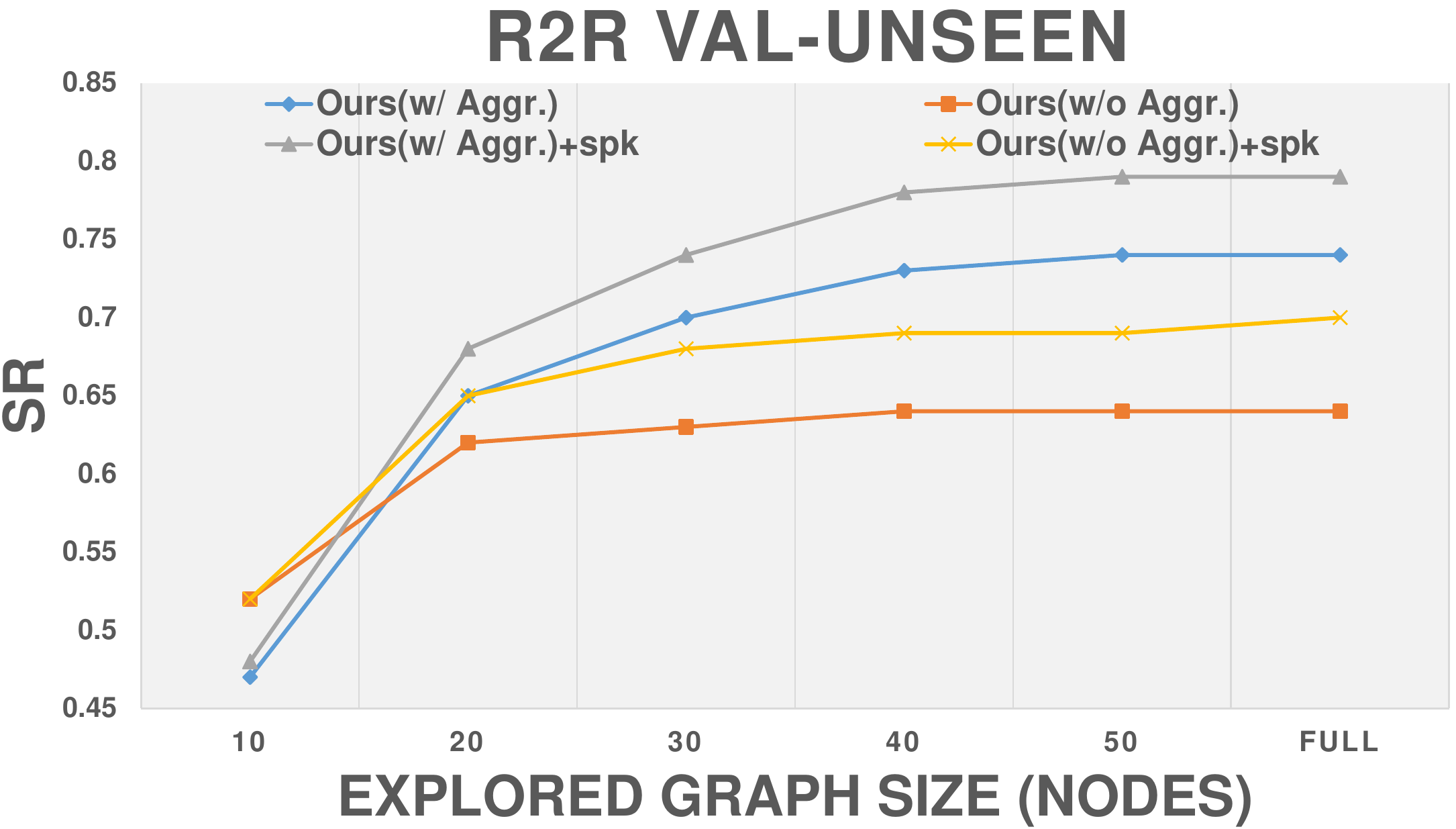}
     \hfill
     \includegraphics[width=0.85\linewidth]{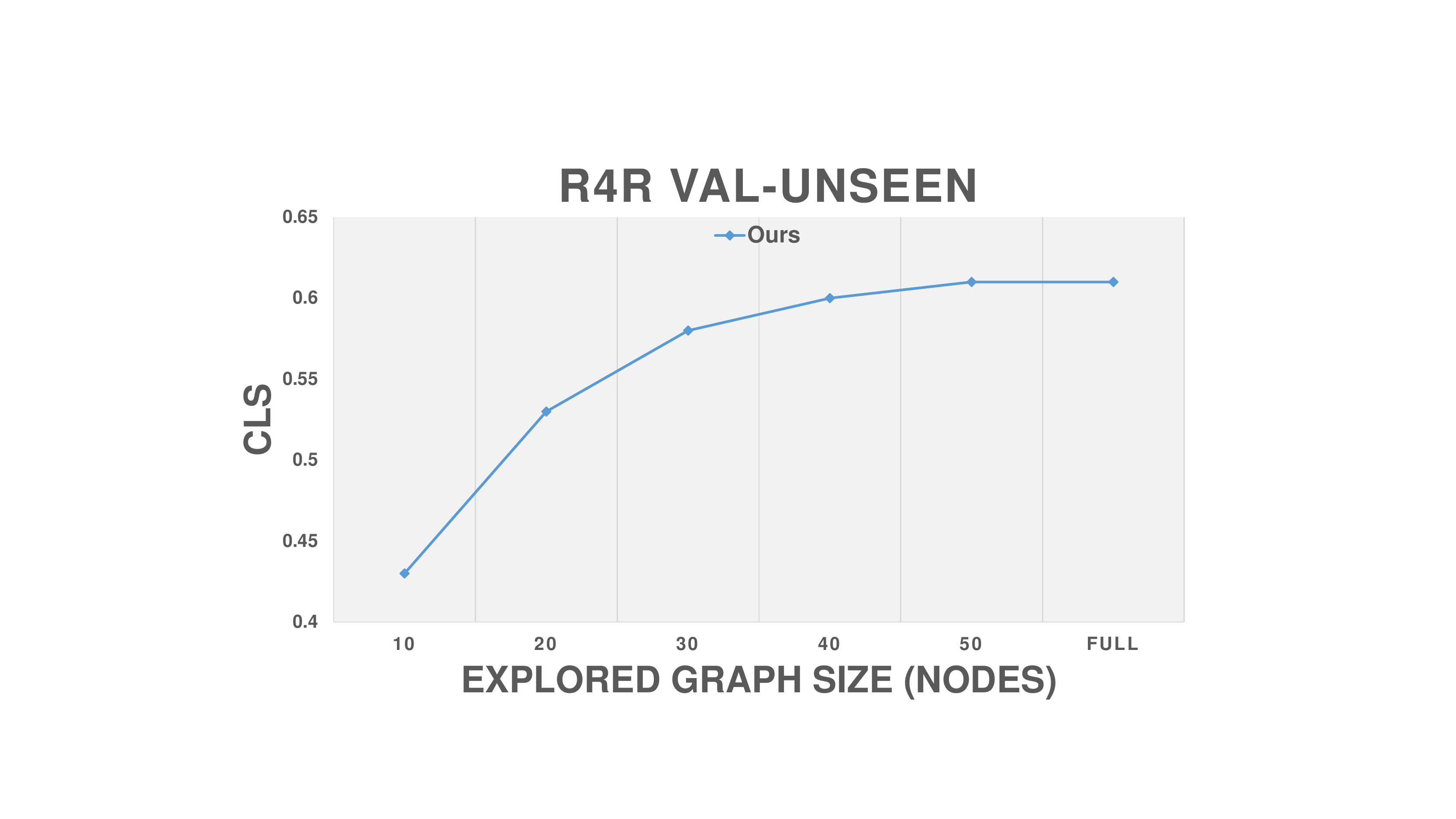}
    \caption{The full list of results by changing the size of the explored-graph on R2R (top) and R4R (bottom). +spk means conducting model ensemble with a Speaker~\cite{FriedHCRAMBSKD18} model.}
    \label{fig:ablate}
\end{figure}

\subsection{Ablation Study of Key Components}\label{ssec:ablate}
To showcase the impact of several key design choices to the final performance, we conduct ablative studies on each of them to provide a comprehensive understanding of our method.

\textbf{Size of the Explored Sub-Graph.} In Section~\ref{ssec:r2r} and \ref{ssec:r4r}, we set the size of the explored sub-graph to 40 nodes by default. Here we provide a full list of results with various sub-graph sizes in Figure~\ref{fig:ablate} for both R2R and R4R. It is clear that by increasing the size of the explored graph, the performance grows better and approaches the result of full map. Empirically, we find that setting the graph-size to 40 (nodes) strikes a good balance between the performance and the cost.   

\textbf{Neighboring Score Aggregation.} The operation of neighboring score aggregation helps bridge the gap between the accuracy and SR, and here we compare the results with and without the aggregation in the first panel of Figure~\ref{fig:ablate} on R2R. Note for R4R, the sequential model could not compute the exact probability of each node being the goal since it needs to sum over all possible paths, so we skip the aggregation directly. From Figure~\ref{fig:ablate} we can see that when the graph-size is small, \emph{e.g.}, 10 nodes, conducting aggregation actually hampers the performance, and when the graph-size grows larger ($\geq 20$), it instead contributes to the result. The reason is that as the topology of the sub-graph is incomplete, the aggregation may either be imprecise or unbalanced. As the graph-size increases, the topology is more close to the full map so the aggregation brings more benefit.

\begin{table}[h!]
  \centering
  \caption{Ablation study of the progress-aware scoring function (top) and the exploration agent (bottom). +spk means conducting model ensemble with a Speaker~\cite{FriedHCRAMBSKD18} model. }
  \label{tab:ablate}
  
  \begin{subtable}{.85\linewidth}{
    \caption{}
    \begin{footnotesize}
    \begin{tabular}{c|ccc}
    \hline
    \multirow{3}{*}{Models} & \multicolumn{3}{c}{R4R} \\
    & \multicolumn{3}{c}{\texttt{val-unseen}} \\ 
    & SR$\uparrow$ & NE$\downarrow$ & \textbf{CLS}$\uparrow$ \\ \hline \hline
    Ours(known map)  MLP & 0.32 & 8.38 & 0.58  \\
    Ours(known map) Aggr$_\text{soft}$ & 0.36 & 7.46 & 0.61  \\ \hline
    Ours(explore@40)  MLP & 0.33 & 7.59 & 0.58  \\
    Ours(explore@40)  Aggr$_\text{soft}$ & 0.36 & 7.02 & 0.60  \\
    \hline
    \end{tabular}
    \end{footnotesize}}
    \end{subtable}%
    \hfill
    \begin{subtable}{.85\linewidth}{
    \caption{}
    \begin{footnotesize}
    \begin{tabular}{c|ccc}
    \hline
    \multirow{3}{*}{Models} & \multicolumn{3}{c}{R2R} \\
    & \multicolumn{3}{c}{\texttt{val-unseen}} \\ 
    & \textbf{SR}$\uparrow$ & NE$\downarrow$ & TL$\downarrow$ \\ \hline \hline
    \multicolumn{4}{c}{beam-search} \\ \hline
    Follower(explore@40) & 0.39 & 6.32 & 636.3 \\
    Follower(explore@40)+spk & 0.58 & 4.32 & 636.6 \\
    Ours(explore@40) & 0.73 & 3.61 & 636.0 \\
    Ours(explore@40)+spk & 0.78 & 3.13 & 636.1 \\ \hline
    
    SSM + Ours & 0.61 & 4.22 & 47.1 \\
    SSM + Ours + spk & 0.63 & 4.11 & 47.0 \\
    \hline
    \multicolumn{4}{c}{single-run} \\ \hline
    SSM & 0.60 & 4.58 & 45.4 \\
    \hline
    \end{tabular}
    \end{footnotesize}}
    \end{subtable}
\end{table}

\textbf{Progress-Aware Scoring Function.} In Section~\ref{ssec:relax}, we propose to replace the MLP scoring function with a progress-aware module to respect the progress of partial paths on R4R. Here we quantitatively characterize its advantage in Table~\ref{tab:ablate}(a). It is clear that using Aggr$_\textbf{soft}$ is indeed superior than MLP, but we also notice that even with the sub-optimal MLP function, the model still out-performs previous methods in Table~\ref{tab:r4r}, demonstrating the effectiveness of utilizing the route prior. 

\textbf{Solely the Exploration Agent.} The explore-and-exploit scheme re-purposes an off-the-shelf VLN agent to build the sub-graph for our model. Here we compare the result with solely the exploration agent (the Follower) , \emph{i.e.} the 2nd stage path-finding is also based on the Follower, to demonstrate that the superiority indeed relies on our model. The results are shown in Table~\ref{tab:ablate}(b). It is clear that solely the Follower model is far worse than our model. Note in Table~\ref{tab:ablate} we count all the exploration trajectory in TL for fair comparison.

\textbf{Combining with Other Exploration-Agent.} Lastly, we show the generalizability of our model by replacing the Follower model with a state-of-the-art single-run agent, SSM~\cite{WangWLXS21}, for exploration. Specifically, we run SSM in the single-run mode for exploration, and base the 2nd stage on it. 
As shown in Table~\ref{tab:ablate}(b), our method could still improve the performance based on purely the single-run explored graph of SSM which is far smaller than that of the beam-search agent, \emph{e.g.} the average size of the explored graph is only about 11 nodes. 

\section{Concluding Remarks}
This work addresses the task of VLN from a previously-ignored aspect, namely the spatial route prior of the navigation scenes. For the most information-rich case of knowing environment maps and admitting shortest-path prior, we transform the route-finding problem into a node-classification task and design a straightforward model for it. Then we relax the route prior to a general case with a sequential-decision model and remove the map-awareness assumption with an explore-and-exploit scheme to fit our model to more realistic situations. Extensive experiments and ablations on R2R and R4R clearly demonstrate the effectiveness of our design. Overall, we hope our work could help inspire new thoughts in VLN to further promote the development.

{\small
\bibliographystyle{ieee_fullname}
\bibliography{egbib}
}

\end{document}